\documentclass{article}

\usepackage{microtype}
\usepackage{graphicx}
\usepackage{subfigure}
\usepackage{booktabs} 
\usepackage{adjustbox}
\usepackage{array}

\usepackage{hyperref}

\usepackage[accepted]{icml2025}

\usepackage{amsmath}
\usepackage{amssymb}
\usepackage{mathtools}
\usepackage{amsthm}
\usepackage{multirow}

\usepackage[capitalize,noabbrev]{cleveref}

\usepackage{setspace}

\newcommand{\explanation}{\mathcal{R}^{t_k}}
\newcommand{\computationgraph}{\mathcal{G}^{t_k}_c}
\newcommand{\inputgraph}{\mathcal{G}^{t_k}}
\newcommand{\df}{$\alpha\text{Fidelity}$}
\newcommand{\deltaf}{$\Delta\text{Fidelity}$}

\theoremstyle{plain}
\newtheorem{theorem}{Theorem}[section]

\theoremstyle{definition}
\newtheorem{definition}[theorem]{Definition}

\theoremstyle{remark}

\usepackage[textsize=tiny]{todonotes}

\icmltitlerunning{XAI for CTDG Spatio-Temporal Models}

\begin{document}

\icmltitle{STX-Search: Explanation Search for Continuous Dynamic Spatio-Temporal Models}



\icmlsetsymbol{equal}{*}

\begin{icmlauthorlist}
\icmlauthor{Saif Anwar}{yyy}
\icmlauthor{Nathan Griffiths}{equal,yyy}
\icmlauthor{Thomas Popham}{equal,yyy}
\icmlauthor{Abhir Bhalerao}{equal,yyy}
\end{icmlauthorlist}

\icmlaffiliation{yyy}{Department of Computer Science, University of Warwick, United Kingdom}

\icmlcorrespondingauthor{Saif Anwar}{saif.anwar@warwick.ac.uk}

\icmlkeywords{Machine Learning, ICML}

\vskip 0.3in



\printAffiliationsAndNotice{\icmlEqualContribution} 

\begin{abstract}
Recent improvements in the expressive power of spatio-temporal models have led to performance gains in many real-world applications, such as traffic forecasting and social network modelling. However, understanding the predictions from a model is crucial to ensure reliability and trustworthiness, particularly for high-risk applications, such as healthcare and transport. Few existing methods are able to generate explanations for models trained on continuous-time dynamic graph data and, of these, the computational complexity and lack of suitable explanation objectives pose challenges. In this paper, we propose \textbf{S}patio-\textbf{T}emporal E\textbf{X}planation \textbf{Search} (STX-Search), a novel method for generating instance-level explanations that is applicable to static and dynamic temporal graph structures. We introduce a novel search strategy and objective function, to find explanations that are highly faithful and interpretable. When compared with existing methods, STX-Search produces explanations of higher fidelity whilst optimising explanation size to maintain interpretability.
\end{abstract}

\section{Introduction}
\label{sec:intro}
Many real-world applications of machine learning involve data with spatial and temporal domains, such as forecasting of traffic flow, weather, and disease spread \cite{Sofi2022,yuanSurveyTrafficPrediction2021}. The spatial aspects of these data can be depicted as graph structures where nodes and edges are used to represent entities and the relationships between them. Graph Neural Networks (GNNs) have been shown to be effective in learning spatial relationships \cite{Zhou_Cui_Hu_Zhang_Yang_Liu_Wang_Li_Sun_2020, WuGraphSurvey}. Spatio-temporal models adopt a fused architecture which combines GNNs with temporal models to learn both spatial and temporal dependencies \cite{Zhao2020, Han2023}. For example, STGCN uses a Graph Convolution Network (GCN) and a Temporal Convolution Network (TCN) to capture the respective dependencies \cite{yuSpatioTemporalGraphConvolutional2018}. 

Spatio-temporal models have been shown to have high expressive power, however the underlying predictive behaviour lacks transparency \cite{Berkani2023, yuanSurveyTrafficPrediction2021}. Explainability methods aim to understand the reasoning for model predictions and may be divided into post-hoc and inherently-interpretable categories. The former includes methods that explain the predictions of an existing trained model, whereas the latter includes models that are designed to explain their own predictions. A case can be made for both categories, however this paper focuses on post-hoc methods \cite{agarwalEvaluatingExplainabilityGraph2023,yuanExplainabilityGraphNeural2022}. 

Many works have been proposed to understand the behaviour of GNNs, such as GNNExplainer \cite{yingGNNExplainerGeneratingExplanations} which attempts to find the most influential subgraph of the input graph for a given prediction, and PGM-Explainer \cite{vuPGMExplainerProbabilisticGraphical2020}, which presents a Bayesian network to indicate the dependencies between the variables in the input graph. Although these methods cannot be directly applied to spatio-temporal models, they have been used as a foundation for developing more suitable explainability techniques \cite{xiaEXPLAININGTEMPORALGRAPH2023,tangExplainableSpatioTemporalGraph2023,heExplainerTemporalGraph2022}. 

Spatio-temporal data can be regarded as either static or dynamic \cite{Yuan_Li_2021}. Static data are those for which the spatial relationships in the graph structure remain constant over time, such as a road network. Dynamic structures are those where the spatial relationships are time-varying, such as social networks or vehicle trajectories \cite{kazemiRepresentationLearningDynamic2020}.  For spatio-temporal models, many explanation methods, such as TGNNExplainer, use search-based methods to find a subset of the input data that is most influential in providing the prediction \cite{xiaEXPLAININGTEMPORALGRAPH2023}. Since the number of possible explanations grows exponentially with the input size, the search must be guided in a computationally efficient manner with appropriate objectives to evaluate the quality of the explanation. Other explanation methods aim to identify important sub-structures within the input, such as TempME, which samples a number of temporal-motifs and assigns importance scores to them \cite{chenTempMEExplainabilityTemporal2023}. Similarly, the number of possible sub-structures grows exponentially and they must be sampled carefully to ensure the most influential motifs are found.

In this paper, we propose a novel method, STX-Search, for generating instance-level explanations, with our contributions summarised as follows:
\begin{itemize}
    \item A computationally efficient search-based method for finding the subset of the input data that is most influential towards the prediction for a specific input data instance made by a black-box spatio-temporal model. Our method is applicable to both static and dynamic data used to make node, graph or edge-level predictions in classification and regression tasks.
    \item A novel search strategy and objective function to quantify explanation quality by balancing fidelity and sparsity of the explanation. We show that our method is able to generate explanations that are both accurate and concise.
\end{itemize}

\section{Background}
\label{sec:background}
\subsection{Spatio-Temporal Data}
\label{sec:spatio-temporal-data}
Spatio-temporal data can be static or dynamic, where the spatial relationships between entities remain constant or change over time. Static data can be represented as a graph $\mathcal{G} = \{V,E, [X_t,...,X_{t+T-1}]\}$ where $V$ is the set of nodes, $E$ is the set of edges, and $X_t$ denotes the features of the nodes and edges at time $t$ \cite{kazemiRepresentationLearningDynamic2020}. The length of the temporal sequence is denoted by $T$. Dynamic graph structures can be further divided into Discrete Time Dynamic Graphs (DTDGs) and Continuous Time Dynamic Graphs (CTDGs) \cite{rossiTemporalGraphNetworks2020}. In DTDGs, the graph structure changes at discrete time intervals, and can be represented as a series of timestamped graphs, $[\mathcal{G}_t,...,\mathcal{G}_{t+T-1}]$. Each graph $\mathcal{G}_{t+i} = [V_{t+i}, E_{t+i}, X_{t+i}]$ is represented by the vertices and edges it contains at time $t+i$, as well as the state of their attributes at time $t+i$, such that the graph occurs $i$ timestamps into the future from $t$. In CTDGs, the structure of the graph changes continuously and is denoted as $\mathcal{G}^{T} = \{\mathcal{S}^{T}, \mathcal{N}^{T}\}$. The spatio-temporal data consists of a continuous sequence of temporal events $\mathcal{S}^{T} = \{e_0,...,e_N\}$ that occur up until but not including time $t+T$, as well as the set of nodes, $\mathcal{N}^{T}$, involved in $\mathcal{S}^T$ \cite{gravinaLongRangePropagation2024}. Each event $e_n \in \mathcal{S}$, is structured as $e_n = \{s_{n}, d_{n}, t_n, att_n\}$ and represents a change to a node or edge entity within the graph. This describes an event occurring at time $t_n$, between nodes $s_{n}$ and $d_{n}$. The vector $att_n$ indicates a change in the attributes or addition/deletion of the entity. In the case of an edge event, $s_{n}$ and $d_{n}$ are the nodes connected by the edge, whereas for a node event, $s_{n}$ is the node that has changed and $d_{n}$ is set to null. Here we note that methods developed for CTDGs can be applied to both DTDGs and static graphs, therefore we will focus on CTDGs to develop a truly model-agnostic explainability approach.

\subsection{Spatio-Temporal Models}
\label{sec:spatio-temporal-models}
Spatio-temporal models take in a graph data structure with both spatial and temporal elements to make predictions for entities in the future based on historic observations. Suppose a prediction is being made for a node or edge, called the target entity, of an event $e_{k} = \{s_{k}, d_{k}, t_k, att_k\}$, called the target event. A model $f(\cdot)$ will take $\mathcal{G}^{t_{k}}$ as input, which is all events and nodes occurring before $e_k$, to make a prediction. The model output $f(\inputgraph)$ will contain predictions for all target entities in the graph at time $t_k$. For example, if the model is predicting the presence of an edge between two nodes at $t_k$, a distribution will be predicted for all possible edges. For regression tasks, the model will output a continuous value for attributes of the target entities. Spatio-temporal models learn the spatial relationships within the data through a process called \emph{message-passing} \cite{waikhomGraphNeuralNetworks2021,velickovicGraphAttentionNetworks2018}, which aggregates neighbourhood information to learn a hidden representation that is used to make predictions. Some works have suggested connecting a node to itself in neighbouring timestamps via \emph{temporal edges} and using the same mechanism to learn the temporal dependencies \cite{rossiTemporalGraphNetworks2020}. However, message-passing focuses on learning local relationships and struggles to capture long-range temporal dependencies. To address this, some methods \cite{xuInductiveRepresentationLearning2020,yuSpatioTemporalGraphConvolutional2018} propose an architecture that learns the spatial relationships as described above, combined with a model that is more suited for capturing the long-term temporal dependencies such as an LSTM or TCN \cite{Yuan_Li_2021}. The variations for fusing the learnt representations of the spatial and temporal dependencies has been researched extensively \cite{yuanSurveyTrafficPrediction2021,longaGraphNeuralNetworks2023}, however, this is outside the scope of this paper.

\subsection{Problem Definition}
\label{sec:problem-definition}
We wish to explain the prediction made by a spatio-temporal model $f(\inputgraph)$ for a specific target event $e_k$. We define an explanation to be a subset $\explanation$ of the input data $\inputgraph$ that is most influential towards the prediction of $e_k$. From the set of all possible explanations, which is the power set $\mathcal{P}(\inputgraph)$, we aim to find the $\explanation$ such that $f(\explanation)[e_k]$ is as close as possible to $f(\inputgraph)[e_k]$. It can be assumed that as events are removed from $\inputgraph$, the prediction for $e_k$ will deviate from the original prediction $f(\inputgraph)[e_k]$. A completely faithful explanation, with maximum fidelity, is one which produces an identical output to the original prediction, an example of which is $\explanation = \inputgraph$. However, this is not useful as it does not provide any insights into the predictive reasoning and does not improve \emph{interpretability}, which we define to be the notion that an end user may understand the explanation. It can be assumed that there is a trade-off between interpretability and fidelity, where a more interpretable explanation will be less faithful and vice versa. We aim to find the subset of the input data that is most influential towards the prediction of the target entity, while balancing the trade-off between interpretability and faithfulness. 

\subsection{Evaluating explanation quality}
In existing explanation methods for spatio-temporal models, an objective function based on mutual information (MI) is used when searching for an explanation \cite{xiaEXPLAININGTEMPORALGRAPH2023,seoSelfExplainableTemporalGraph2024}. MI is a measure of how much information a random variable contains regarding another and can be used to quantify the shared information between the explanation and the prediction \cite{Taverniers2021}. This may be appropriate for classification tasks where the model predicts a distribution over the target space, however for regression tasks where the model predicts a continuous value, MI is not suitable. Also, MI has been shown to be inconsistent with explanation fidelity, especially when the number of events in the graph is large \cite{rongEfficientGNNExplanation2023}. In existing works for explaining GNNs, Fidelity$^+$ and Fidelity$^-$ have been proposed as metrics for evaluating explanation quality, as shown in Equations \ref{eq:fidelity+} and \ref{eq:fidelity-} \cite{liuInterpretabilityGraphNeural2022,Yuan_Li_2021,zhangGStarXExplainingGraph2024}.
\begin{equation}
    \label{eq:fidelity+}
    \text {Fidelity}^{+}(e_k, \explanation)= f(\inputgraph)[e_k]-f(\inputgraph \backslash \explanation)[e_k]
\end{equation}
\begin{equation}
    \label{eq:fidelity-}
    \text {Fidelity}^{-}(e_k, \explanation)= f(\inputgraph)[e_k]-f(\explanation)[e_k]
\end{equation}
Rong et al. combine these to form a single 
\deltaf
metric, shown in Equation \ref{eq:delta-fidelity}. This measures the difference between the prediction for $e_k$ made by the model using the important nodes, contained within the explanation $\explanation$, and the prediction made by the model once important nodes are removed from the original input. 
\begin{equation} 
    \label{eq:delta-fidelity}
    \begin{split}
        \Delta\text{Fidelity}(e_k, \explanation) & = \text{Fidelity}^{+}(e_k, \explanation) - \\ & \text{Fidelity}^{-}(e_k, \explanation)
    \end{split}
\end{equation}
As mentioned earlier, explanations should be interpretable as well as faithful. Existing works propose a \emph{Sparsity} metric which measures the explanation graph size as a proportion of the input graph size \cite{Yuan_Li_2021}. This is currently present only in methods for explaining GNNs but can be simply extended to spatio-temporal models and calculated as $(1-|\explanation|/|\inputgraph|)$.

\subsection{Related Work}
\label{sec:related-work}
Of the existing post-hoc methods for explaining spatio-temporal models, TGNNExplainer \cite{xiaEXPLAININGTEMPORALGRAPH2023}, TempME \cite{chenTempMEExplainabilityTemporal2023} and TempME \cite{chenTempMEExplainabilityTemporal2023} are the only methods that can generate explanations for CTDGs. TGNNExplainer is presented as a method which explains models that predict the occurence of edges. It does this through an explorer-navigator framework where a navigator MLP learns importance scores for each event in the input graph $\inputgraph$, towards the prediction of the target event $e_k$. These importance scores are then used by the MCTS explorer to find the most influential subset of the input graph by removing events from a set of candidate events. An explanation is found once the size of the subset is smaller than some defined threshold. Explanation quality is evaluated throughout the search using a metric based on MI between predictions from the explanation and the original input graph, which has been shown to be inappropriate for evaluating explanations \cite{rongEfficientGNNExplanation2023}. Aside from this, TGNNExplainer does not lend itself to regression tasks where MI cannot be calculated. More importantly, MCTS considers the ordering of event removals to impact the outcome of the search. In the case of an explanation search, the ordering of event removal should not matter. Therefore multiple branches of the search can lead to the same outcome which is computationally expensive, particularly for large graphs. To address this issue, TGNNExplainer initialises the search with a reduced set of candidate events which is defined to be the N most recent events in the input graph. This is an inappropriate assumption as the most recent events may not be the most influential towards the prediction of the target event. 
\begin{figure}[!t]
    \centering
    \includegraphics[width=0.4\textwidth]{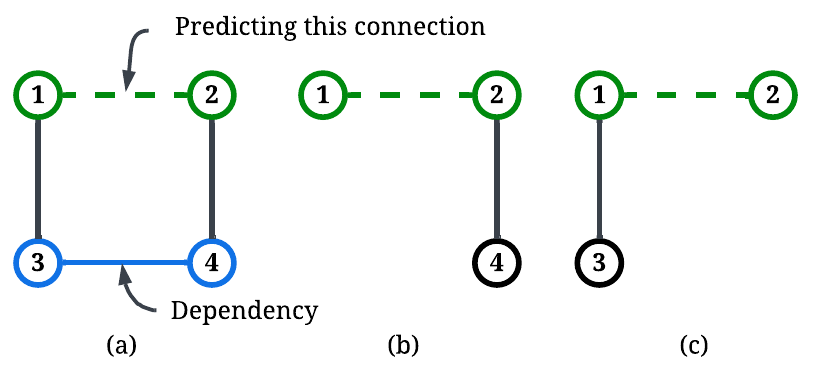}
    \caption{An illustration showcasing the dependency between 2 users within a social network setting. The information regarding the connection between users 3 and 4 may be crucial in predicting the state of the connection between users 1 and 2. }
    \label{fig:dep-def}
\end{figure}
TempME extracts small temporal-motifs from the input graph that are most influential towards the prediction of the target event \cite{chenTempMEExplainabilityTemporal2023}. Since the number of possible motifs grows exponentially with the input graph size,  a motif sampling algorithm is used to generate some set of candidate motifs. These are then assigned importance scores using an approach founded in information-bottleneck (IB) theory \cite{alemiDeepVariationalInformation2019}. IB is rooted in MI with a regularisation component attached to it and, as discussed, an explanation with high MI is not necessarily a faithful one \cite{rongEfficientGNNExplanation2023}. The sampling approach employed by TempME does not guarantee that the most influential motifs are found. Events may have dependencies between them, where the impact of an event within a motif may be dependent on the presence of another event in the motif. If this consideration is not made by the motif sampling procedure, then important events may be missed.
\begin{definition}
\label{def:event-dependencies}
Two or more events are said to have a \emph{dependency} between them if the impact of one event on the prediction of the target event is dependent on the presence of another event. For example, Figure \ref{fig:dep-def} illustrates a simple example of a social network where the presence of a connection between users 1 and 2 is being predicted. In such a scenario, each users connections, and their neighbourhoods, may be of importance. The information gained from users 3 and 4, as in configuration (a), would lead us to believe that the connection between users 1 and 2 is more likely to exist. However, if we only had information regarding one of users 3 or 4, as in configurations (b) and (c), the connection between users 1 and 2 may not be as likely. We say there is a dependency between users 3 and 4 since the information to be gained by one user, ergo its importance towards the prediction, is dependent on the presence of the other.
\end{definition}
In both TGNNExplainer and TempME, the interpretability of the explanation is defined using an explanation size threshold. A fixed explanation size may not be suitable across all instances within a given dataset since individual instances may have different numbers of spatial and temporal neighbours influencing them.
\section{Methodology}
\label{sec:methodology}
In this section, we propose a method for finding a subset of events $\explanation$ from the input graph $\inputgraph$, which are used to explain the prediction made by the base model $f(\inputgraph)$ for a specified target event $e_k$. Following existing works \cite{xiaEXPLAININGTEMPORALGRAPH2023}, a search-based method is employed to find the most influential subset of the input graph. 

\subsection{Search Space}
As mentioned in Section \ref{sec:related-work}, the number of possible explanations is $\mathcal{P}(\inputgraph)$, which grows exponentially with the input graph size. Although we do not make assumptions regarding the internals of the model architecture, it is appropriate to make the assumption that spatial relationships are inferred through message-passing \cite{Li_Yu_Liu_Zhang_Gong_Zhao_2023}. Therefore, the information which contributes to the prediction of a node in the spatial domain is restricted to an L-hop neighbourhood, where L is the number of message-passing layers in the model. The model may capture long-range temporal dependencies, and all spatially relevant events across the entire input temporal window may contribute to the prediction of a target event. Considering these assumptions, we can restrict the search space to only include the events which contribute to the prediction of the target event, known as the computation graph $\computationgraph$. Although this reduces the complexity of the search, the search space is still exponentially large for target events with a large number of spatially relevant events. To ensure a solution is found in a reasonable time existing methods, such as TGNNExplainer, apply a temporal threshold to $\computationgraph$ to only consider the N most recent events (25 by default) out of the events in the input temporal range \cite{xiaEXPLAININGTEMPORALGRAPH2023}. Since the temporal component of the base model can often learn long-range temporal dependencies, all events within the temporal range of the input should be considered for inclusion in the explanation. Therefore a computationally efficient search strategy is required, that considers all events that may contribute to the prediction for the target event $e_k$, i.e., all events in $\computationgraph$.
\subsection{Simulated Annealing Explanation Search}
We propose a simulated annealing strategy to find $\explanation \subseteq \computationgraph$. Simulated annealing randomly generates an initial solution, in this case an explanation $\explanation$, and iteratively perturbs the solution to find a better one. The probability $P(\mathcal{R}^{t_{k'}})$ of accepting the perturbation, $\mathcal{R}^{k'}$, to replace the current solution, $\explanation$, is calculated using the following policy, where $l$ is the objective function for the search and $T$ is the temperature.
\[
    P(\mathcal{R}^{t_{k'}}) =  
\begin{cases}
    1, & \text{if } l(\mathcal{R}^{t_{k'}}) \leq l(\explanation)\\
    e^{\frac{-|l(\mathcal{R}^{t_{k'}}) - l(\explanation)|}{T}}, & \text{otherwise}
\end{cases}
\]
Perturbations are generated by selecting a random event from $\explanation$ and replacing it with a random event from $\computationgraph \backslash \explanation$. The temperature is reduced at each iteration to reduce the probability of accepting a worse solution. In simulated annealing, it is important to occasionally accept worse solutions to escape local optima of the objective function. When generating explanations, this can be understood for cases where an event in $\explanation$ only improves the fidelity and is classed as important given the presence of another event in $\explanation$. If only one of the dependent events is included in the explanation, we may still want to accept it in case the other event is included later on. As the temperature decreases, worse solutions are less likely to be accepted since it is expected that a large portion of the search space will already have been explored and an optimal solution is near. 
\subsection{Search Objective}
We propose a novel strategy which aims to optimise the fidelity and size of the explanation in a multi-stage search. 
\deltaf
as shown in Equation \ref{eq:delta-fidelity}, is a metric used in existing works to evaluate explanation quality \cite{rongEfficientGNNExplanation2023,heExplainerTemporalGraph2022}. It is a measure of the difference in absolute error of predictions made by the base model using events presented in the explanation, Fidelity$^-$, and all events that are not included in the explanation, Fidelity$^+$, which are supposedly unimportant. 
\begin{figure}
    \centering
    \includegraphics[width=0.4\textwidth]{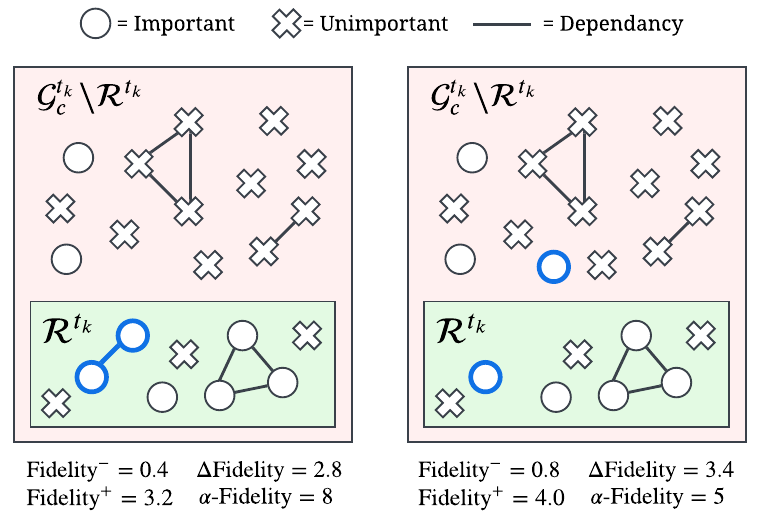}
    \caption{Two examples of an explanation, $ \explanation $, and its complement, $\computationgraph \backslash \explanation$, which when combined give the computation graph. Both explanations contain a number of important events. The explanation on the left contains two groups of dependent events, with one of them highlighted in blue. The explanation on the right no longer contains one of the dependent events in the highlighted pairing, which is now part of the explanation complement, $\computationgraph \backslash \explanation$. The explanation with the dependent events grouped together is more faithful to the base model prediction, and has a lower Fidelity$^-$. However, the less faithful explanation on the right incorrectly results in a higher $\Delta$Fidelity.}
    \label{fig:dependancies}
    \vspace{-0.5cm}
\end{figure}
However, \deltaf may not suitably correlate with explanation performance in all cases. First we consider the case where the explanation gives a prediction with low Fidelity$^-$ and the explanation complement, $\computationgraph \backslash \explanation$, gives a prediction with high Fidelity$^+$. In this case, \deltaf will be high, and appropriately translates to explanation performance. It may not always be the case however that a decrease in Fidelity$^-$ leads to an increase in Fidelity$^+$ since the importance of an event may be dependent on the presence of another event being in the explanation, as decribed in Definition \ref{def:event-dependencies}. For example, the left side of Figure \ref{fig:dependancies} shows important events with dependencies contained within an explanation. Removing one of these dependent events from the explanation whilst leaving the other, as shown in the subsequent explanation, would cause detriment to the explanation performance and increase Fidelity$^-$. Since this event is not classed as important without its dependent partner, Fidelity$^+$ may also increase, and disproportionately so such that \deltaf may increase overall. Although, we now have an explanation with higher error, the \deltaf metric would suggest that the explanation is better. To address this, we propose an alternative metric, \df, that takes the following form.
\begin{equation}
    \label{eq:adjusted-delta-fidelity}
        \alpha\text{Fidelity}(e_k, \explanation) = \frac{\text{Fidelity}^{+}(e_k, \explanation)}{\text{Fidelity}^{-}(e_k, \explanation)}
\end{equation}
This adjustment ensures that for the \df to improve either Fidelity$^-$ must decrease more significantly than Fidelity$^+$, or the inverse. The values used to calculate Fidelity$^-$ and Fidelity$^+$ vary for different types of predictive tasks, which is further detailed in Appendix \ref{app:multitask-explanations}. It must be noted that it is not sufficient to only observe the change in \df when determining the quality of the explanation, since it does not provide any information directly regarding the explanation's prediction compared to that of the base model. Rather, it indicates whether there are proportionately more important events in $\explanation$ than in $\computationgraph \backslash \explanation$. Therefore, absolute error of the explanation prediction, when compared to the original base model prediction, must also be considered within the scoring function used to quantify explanation performance. To ensure that the explanation is interpretable, we insert a regularisation term which penalises larger explanations. Combining these components, we propose the following objective function.
\begin{equation}
    \label{eq:objective-function}
    \begin{split}
    l(\explanation) = \epsilon \cdot \lvert f(\computationgraph) - f(\explanation) \rvert + \\ \gamma \cdot \alpha\text{-Fidelity}(e_k, \explanation)- \lambda \cdot \text{Sparsity}(\explanation)
    \end{split}
\end{equation}
The parameters $\epsilon$, $\gamma$ and $\lambda$ are hyperparameters which dictate the importance of the fidelity, sparsity and \df components of the objective function. These are controlled within the different stages of the search process. In the first stage, $\gamma$ and $\lambda$ are set to 0 and $\epsilon=1$ so that an initial explanation is found which is highly faithful to the base model prediction. In the second stage, $\gamma$ is also set to 1 and $\lambda$ remains at 0. Although many important events will be found during the first search stage, the second stage ensures that important events remaining in $\computationgraph \backslash \explanation$ are also included in the explanation. Also in this stage, events that may not impact the error of the explanation but increase the error of the complement should be removed since they do not have a positive impact. In the final stage, $\lambda$ is set to some value between 0 and 1. This stage aims to reduce the size of the explanation by removing unimportant events whilst maintaining a high fidelity. The value of $\lambda$ is used to dictate whether a more interpretable, i.e., more sparse, explanation is preferred or a more accurate and faithful one. We leave this to be defined by the  user based on the application context, for example a more faithful explanation may be preferred in higher-risk use cases.

\section{Evaluation Methodology}

\textbf{Datasets:} We evaluate the performance of STX-Search on the real-world Reddit and Wikipedia datasets \cite{kumar2019predicting}, which are commonly used to model temporal interaction networks. The Wikipedia dataset contains 1 month of page edits between nodes of 1000 users and 8227 pages. Edges are defined using interactions, or events, containing 172 text based features that describe a page being edited by a user. Similarly, the Reddit dataset contains 1 month of interactions between 10,000 users and the 984 most popular subreddits, where interactions are posts made by users on subreddits and are described using the same text features. For both datasets, the label for each interaction is whether a user was banned from the respective platform. Further details regarding the datasets are provided in Appendix \ref{app:datasets}.

\subsection{Base Models}
To evaluate the performance of STX-Search, we explain predictions from two state-of-the-art spatio-temporal models, TGAT \cite{velickovicGraphAttentionNetworks2018} and TGN \cite{rossiTemporalGraphNetworks2020}. TGAT is a model that uses a Graph Attention Network (GAT) to learn spatial relationships and a Temporal Convolution Network (TCN) to learn temporal dependencies. TGN uses a Graph Convolution Network (GCN) to learn spatial relationships and a Long Short-Term Memory (LSTM) to learn temporal dependencies. Although the model performance on the data is not of significance in the explanation task, we ensure that the models are well trained with further performance details provided in Appendix \ref{app:models}.

\textbf{Baselines:} We compare the performance of STX-Search with, to the best of our knowledge, the only two other existing methods for explaining continuous dynamic spatio-temporal models, namely TGNNExplainer and Temp-ME. TGNNExplainer \cite{xiaEXPLAININGTEMPORALGRAPH2023} is a search-based method which finds the most influential subset of the input graph. The search is carried out using MCTS with a 2-layer MLP trained to learn the importance of events, which is used to navigate the search. Temp-ME \cite{chenTempMEExplainabilityTemporal2023} samples a number of possible motifs within the data, where a motif is defined as 3 nodes that are temporally connected. A novel sampling algorithm based on information bottleneck principles is used to generate a number of motifs, where an MLP is trained to learn the importance scores for the motifs, which is presented as an explanation. Both the navigator in TGNNExplainer and the importance scoring MLP in TempME are trained for 100 epochs, as suggested by their literature, with other hyperparameters left as the default values. To maintain consistency with previous literature, we also compare the performance of STX-Search with a version of PGExplainer \cite{Luo_Cheng_Xu_Yu_Zong_Chen_Zhang_2020} that has been adjusted for spatio-temporal models. In its original form, PGExplainer is applied to GNN models to assign importance scores to features of the underlying data structure, such as edges or nodes. An adjusted version proposed in \cite{xiaEXPLAININGTEMPORALGRAPH2023} is used to instead to assign importance scores to individual events within the input graph.

\subsection{Evaluation Metrics}
To evaluate the performance of explanations, we measure the Mean Absolute Error (MAE) between the prediction made by the base model using the set of events found by the explanation, $\explanation$, and the base model prediction using the full computation graph $\computationgraph$. When comparing predictions for classification tasks, we observe the model prediction before any activation function is applied in the final layer, as done in previous works \cite{xiaEXPLAININGTEMPORALGRAPH2023}. This is to remove an element of known behaviour from the model and requires the explanation method to more accurately capture the black-box behaviour. For both STX-Search and TGNNExplainer, predictions are generated by masking out all events not included in the explanation. Since Temp-ME assigns importance scores to motifs, we gather the events contained within the motifs with the highest importance scores and use these to generate predictions by masking out all other events.

We also use the adjusted \df measure proposed in Equation \ref{eq:adjusted-delta-fidelity} to indicated the proportion of important events included in the explanation compared to those left out. Fidelity$^-$ is calculated by masking out all events not included in the explanation, $\computationgraph \backslash \explanation$, whilst Fidelity$^+$ is calculated by masking out events in the explanation.

Although many prior works report sparsity as a measure of interpretability, we do not feel it is appropriate since it is relative to the computation graph size, which may vary. If $\computationgraph$ contains 400 events, a sparsity of 0.2 would still contain 80 events and may not be so interpretable. Instead, we propose to measure the absolute explanation size as the number of events included in $\explanation$.

\subsection{Experiments}
For each dataset, 100 instances are selected at random for which explanations will be generated. For each instance, an explanation of various sizes is generated. By default, TGNNExplainer only considers 25 events in the search space to improve the computation time of the MCTS. We remove this restriction to include all events in the computation graph, which on average contains $\sim$250 events for the evaluated datasets and models. Since this significantly impacts computation time, we limit the number of rollouts to 100. Although this is a reduction from the default 500, we find that improving the search space is more beneficial than increasing the number of roll-outs and allows TGNNExplainer to be more competitive with STX-Search. The MAE and average \df are calculated over the 100 instances for each explanation size.

For STX-Search, each search stage is run for 500 iterations. First, we generate explanations for the requested sizes using the first 2 search stages only, i.e., no sparsity reduction, to allow for a fair comparison with TGNNExplainer and Temp-ME. We then generate explanations for the same instances using all 3 search stages to allow the search to automatically find the appropriate explanation size. We test this for a range of $\lambda$ values. The initial temperature is set to 1 with a cooling rate of 0.99. All experiments are run on a Ryzen 5 3600 CPU@3.6GHz and a RTX 3070Ti GPU.

To encourage reproducibility and further research, our code is available on Github at xxxxx and will also be integrated into the popular open source spatio-temporal forecasting library, LibCity \cite{Jiang2024}, for easy access.
\section{Results \& Discussion}
\begin{table*}[!t]
\centering
\caption{The best average MAE and \df achieved by each method out of all tested explanation sizes. The explanation size that achieved the best result is also shown in brackets. The best performing method for each metric is shown in bold whilst the second best performing method is shown as underlined.}
\label{tab:results}
\begin{tabular}{@{}llccccccc@{}}
\toprule
\multirow{2}{*}{\textbf{Dataset}} & \multirow{2}{*}{\textbf{Model}} & \multirow{2}{*}{\textbf{Metric}}& \textbf{PGExplainer} & \textbf{Temp-ME} & \textbf{TGNNExplainer} & \multicolumn{2}{c}{\textbf{STX-Search}} \\
 & & & & & & \textbf{(Fixed Size)} & \textbf{($\lambda=0.1$)} \\
\midrule
\multirow{4}{*}{Wikipedia} & \multirow{2}{*}{TGAT} & MAE & 0.2784 (100) & 0.2185 (100) & 0.2328 (100) & \underline{0.0566 (80)} & \textbf{0.0006 (37)} \\
 &  & $\alpha$Fid & 945.2 (100) & 4.1 (30) & 5898.5 (100) & \textbf{23359.5 (90)} & \underline{9234.1 (37)} \\
 & \multirow{2}{*}{TGN} & MAE & 0.1939 (100) & 0.3303 (10) & 0.5064 (100) & \underline{0.0944 (80)} & \textbf{0.0020 (17)} \\
 &  & $\alpha$Fid & 524.9 (100) & 1.1 (20) & 100.8 (90) & \textbf{95978.3 (70)} & \underline{1763.4 (17)} \\
\midrule
\multirow{4}{*}{Reddit} & \multirow{2}{*}{TGAT} & MAE & 3.7366 (100) & \underline{0.3120 (30)} & 3.6863 (100) & 1.0495 (100) & \textbf{0.0001 (33)} \\
 &  & $\alpha$Fid & 2.0 (70) & 1.4 (40) & 69.3 (90) & \textbf{17851.0 (100)} & \underline{7677.9 (33)} \\
 & \multirow{2}{*}{TGN} & MAE & 2.0655 (100) & \underline{0.1711 (80)} & 1.3693 (100) & 1.0495 (100) & \textbf{0.0003 (24)} \\
 &  & $\alpha$Fid & 2.8 (100) & 1.1 (10) & 49.9 (60) & \underline{10938.7 (100)} & \textbf{53242.3 (24)} \\
\bottomrule
\end{tabular}
\end{table*}

\begin{figure*}[!ht]
    \centering
    \includegraphics[width=\textwidth]{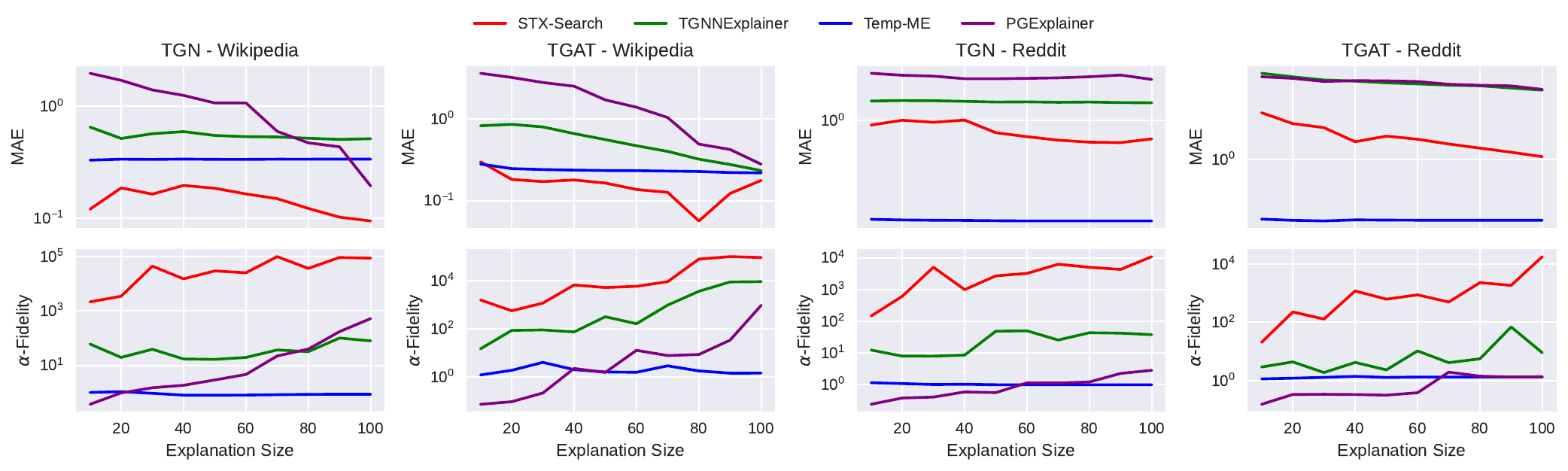}
    \caption{A comparison of the average MAE and \df achieved by each method when generating explanations of different sizes for 100 random instances from each dataset.}
    \label{fig:results}
\end{figure*}

In this section, we present the results of the described experiments and discuss the performance of STX-Search in comparison to existing methods. We also discuss the impact of the $\lambda$ hyperparameter on the explanation size and fidelity. Table \ref{tab:results} shows the best average MAE and {\df} achieved by each method over all fixed explanation sizes, alongside the explanation size that achieved the score. The average MAE and {\df} achieved by STX-Search whilst automatically finding the best explanation size is also reported. 

It is of immediate note that STX-Search significantly outperforms all other methods across all datasets and models for both metrics. This means that STX-Search finds a set of explanation events that is much more likely to contain the events used by the base model to make its prediction, as evident by the extremely low MAE. Not only this, but the proportion of important events in the explanation compared to those left out, as indicated by the $\alpha$-Fidelity, is also much higher than the other methods. The performance of STX-Search is consistent across all datasets and models, highlighting its capability as a robust method for generating explanations for spatio-temporal models.

Figure \ref{fig:results} shows the average MAE and \df achieved across the different explanation sizes. Some cases of low MAE and high \df may be attributed to the large explanation sizes. This is not particularly impressive since out of ~250 events in the computation graph, an explanation of 100 events is not so interpretable. However, the most performant explanations from the baseline methods, which are of size 100, are outperformed by the least performant, and smallest, explanations from STX-Search across all datasets and models. This shows that STX-Search is able to generate explanations that are both faithful and interpretable.

Generally, as explanation size increases, the performance of the explanations generated using TGNNExplainer and STX-Search also increases. Although this is less prominent for TGNNExplainer, it is more significant for STX-Search. Temp-ME on the other hand experiences very little variation in explanation performance as explanation size varies. Explanations generated by Temp-ME contain importance scores generated for a number of motifs. The nodes involved may occur in multiple motifs. In cases where a node is involved in both a motif with high importance and one with low importance, the overall impact of the node is diluted. Since motifs are not sampled with the dependencies between events in mind, for a large number of generated motifs, an important node may often be grouped with unimportant nodes. If the motif containing the important node is then given a lower score, it is unlikely to appear in the final explanation. This dilution of importance may be the reason for the consistent performance of Temp-ME across all explanation sizes, since the random grouping of nodes in motifs leads to an even distribution of importance scores across all nodes. This leads to an even distribution of unimportant and important events in both the explanation and its complement, which may also be the reason for the consistently low \df values. If the performance of $\explanation$ and $\computationgraph \backslash \explanation$ are similar, the \df value will be low, as is often the case with Temp-ME. Although the MAE achieved by Temp-ME on the Reddit dataset is second lowest, the consistently low \df values indicate that this is purely by chance and that the complement of the explanation would be similarly performant. Therefore, no meaningful information can be deduced from the explanation.

As mentioned above, although larger explanations are often more performant, since they contain more of the original input data, it is not always the case. If the explanation contains extra events that have dependencies, but without their dependent partner, it may damage the performance of the explanation. In the case of some larger explanations, the situation may arise that the most dominant important events have been included within the explanation, but there is not enough room to include the remaining important events and their dependencies. This may lead to worse explanation performance compared to if the explanation was smaller. This is evident in the performance of STX-Search for the Wikipedia dataset when explaining the TGN model. The best performing explanation size is 80, which is not the largest. It may not always be possible to know how many events should be included in the explanation. When using all three stages of the search, STX-Search is able to automatically find the best explanation size. This is shown by the performance of STX-Search when using all three search stages with a lambda value of 0.1. The explanations outperform all baseline methods and are also comparable in performance to the best performing fixed size STX-Search explanations whilst being of a more interpretable size.
\begin{figure}
    \centering
    \includegraphics[width=0.48\textwidth]{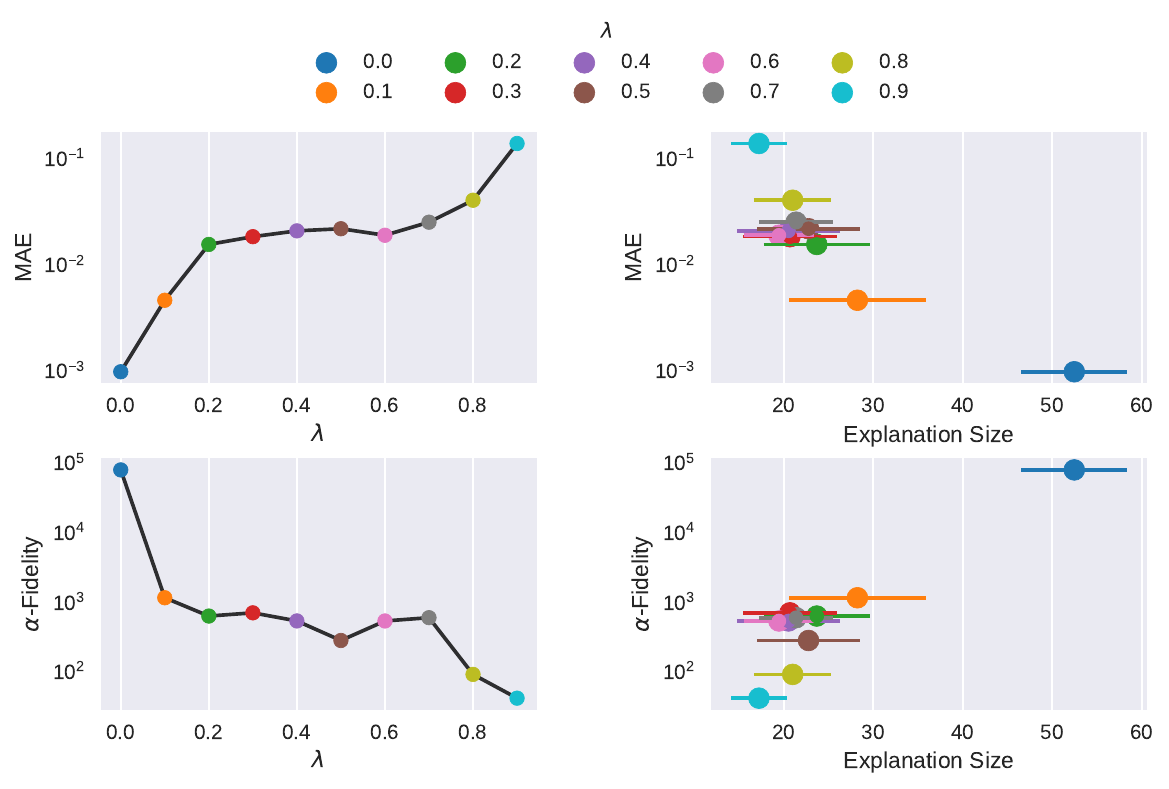}
    \caption{A comparison of average explanation MAE and \df achieved by STX-Search when performing a multi-stage search to automatically find the best explanation size for 100 random instances from the Wikipedia dataset to explain a TGN base model using different $\lambda$ values in the search objective function. The distribution of MAE and \df against explanation size is also shown.}
    \label{fig:lambda}
    \vspace{-0.5cm}
\end{figure}
Figure \ref{fig:lambda} shows the effect of the $\lambda$ hyperparameter on the explanation size and fidelity. The value of $\lambda$ influences the trade-off between explanation size and fidelity. It is assumed that a higher value for $\lambda$ favours a more interpretable explanation. The right side of Figure \ref{fig:lambda} shows the distributions of MAE and \df against the average size of explanations generated using different $\lambda$ values. It can be noticed that for all values of $\lambda$, except $\lambda=0$ where there is no penalty for the explanation size, there is a significant overlap between the range of explanation sizes produced in each case. This is to be expected since STX-Search does not aim for a specific explanation size, and instead aims to find a high performing explanation, and then only reduce the size if it is appropriate to do so. This ensures that we do not encounter the case where explanations are made more interpretable whilst overly sacrificing fidelity. Although the requirement of defining a hyperparameter may be seen as a disadvantage, it is necessary to allow the user to define the level of interpretability required for the explanation. STX-Search then produces explanations of significantly lower error and of much smaller size.

\section{Conclusion}
In this paper, we presented STX-Search, a novel search-based method for generating explanations for continuous dynamic spatio-temporal models. We proposed a novel objective function for the search that allows an explanation to be highly faithful to the behaviour of the base model being explained, whilst maintaining a high level of interpretability by only including neccesary information within the explanation. We compared the performance of STX-Search with, to the best of our knowledge, all other methods within the same application domain across two real-world datasets. We found that STX-Search significantly outperforms all other methods across all test scenarios. We also showed that STX-Search is able to automatically find the best explanation size for a given instance, which is a significant advantage over existing methods. In our future work, we aim to develop a framework for generating a synthetic dataset to test the performance of explanation methods for spatio-temporal models. We aim to control the spatial and temporal dependencies between events such that we may have ground truth explanations that can be evaluated against.

\bibliography{icml2025refs}
\bibliographystyle{icml2025}

\newpage
\appendix
\onecolumn
\section{Multi-Task Explanations}
\label{app:multitask-explanations}
Existing works generating explanations for spatio-temporal models focus solely on link-prediction tasks, such that they explain the logit prediction for whether a link will appear between two nodes. We instead propose a generalised algorithm that is applicable to all predictive tasks with the appropriate adjustments to the objective function required in each case outlined below.
\begin{itemize}
\item \textbf{Node/Edge-Level Classification: }When predicting the presence of a node or edge, the fidelity is calculated using the logit for the target events true class. When classifying nodes or edges out of multiple classes, the fidelity is calculated using the sum of absolute errors of logits over all classes. 
\item \textbf{Graph-Level Classification: }Evaluated by calculating the fidelity over the sum of absolute error for logits over all classes but averaged over all nodes in the graph.
\item \textbf{Node/Link-Level Regression: }Fidelity is calculated using the absolute error between the predicted and true value of the target event.
\item \textbf{Graph-Level Regression: }Fidelity is calculated using the absolute error between the predicted and true value of the target event but averaged over all nodes in the graph.
\end{itemize}

\section{Datasets}
\label{app:datasets}
\textbf{Wikipedia: }This dataset contains edits made to Wikipedia pages within a one-month period \cite{kumar2019predicting}. There are a total of 9227 nodes where nodes are either pages or users. There are 1000 users and 8227 pages with a total of 157,474 interactions describing edits. Each interaction event contains 172 LIWC text based features \cite{Shetty2004TheEE}. The task of our experiments is to predict whether an interaction will lead to the user who made the edit being banned. Until a user is banned, the label is '0', whilst their last interaction has the label '1'. For users that are not ever banned, their label remains '0'. Out of all interactions, there are 217 positive labels (0.14\%). 

\textbf{Reddit: }This dataset contains posts made to 984 most popular subreddits by the top 10,000 users \cite{kumar2019predicting}. The dataset contains 672,447 interaction events describing posts. Similarly, each interaction contains 172 LIWC text base features \cite{Shetty2004TheEE}. The task is the same as the Wikipedia dataset described above where we predict whether a user will be banned or not. There are a total of 372 positive labels (0.05\%).
\section{Models}
\label{app:models}
We use 2 state-of-the-art spatio-temporal models as base models for which we will generate explanations. 
\textbf{TGN: }Temporal-Graph Network (TGN) \cite{rossiTemporalGraphNetworks2020} is a model that is described using the encoder-decoder framework described in \cite{kazemiRepresentationLearningDynamic2020}. An encoder takes a dynamic graph and generates hidden embeddings whilst the decoder performs specified predictive tasks using these embeddings. TGN proposes a novel encoder architecture that is applicable to continuous-time dynamic graphs that takes in a sequence of events to generate hidden embeddings.

\textbf{TGAT: }Temporal-Graph Attention Network (TGAT) \cite{xuInductiveRepresentationLearning2020} uses a temporal self-attention mechanism using a novel attention layer in the encoder of temporal graph networks within the same encoder-decoder framework proposed in \cite{kazemiRepresentationLearningDynamic2020}.

We train both models using the same hyperparameters for both the Wikipedia and Reddit datasets. Both models generate their neighbourhoods using 20 neighbours in each message passing layer with a total of 2 layers. Each model is trained for 100 epochs to minimse the Binary Cross Entropy Loss with a learning rate of 0.0001 and batch size of 512. For each dataset, 70\% of the data is used for training whilst 15\% is reserved for validation and testing. The TGN model achieves a average precision score of 96.8 and 97.2 on the Wikipedia and Reddit datasets respectively, whilst the TGAT model achieves scores of 93.4 and 96.5.

\end{document}